\documentclass[sigchi]{acmart}

\usepackage{booktabs} 
\usepackage{epstopdf}
\usepackage{tabularx}
\usepackage{multirow}
\pdfmapfile{+txfonts.map}
\usepackage{caption}
\usepackage{subcaption}
\def\BibTeX{{\rm B\kern-.05em{\sc i\kern-.025em b}\kern-.08emT\kern-.1667em\lower.7ex\hbox{E}\kern-.125emX}}
    
\copyrightyear{2019}
\acmYear{2019}
\setcopyright{acmlicensed}
\acmConference[RecSys '19]{RecSys '19: ACM Conference on Recommender Systems}{Sep, 2019}{Copenhagen, Denmark}
\acmPrice{15.00}
\acmDOI{10.1145/1122445.1122456}
\acmISBN{978-1-4503-9999-9/18/06}

\begin{document}

%
\title{Latent Multi-Criteria Ratings for Recommendations}

%
\author{Pan Li}
\affiliation{%
  \institution{New York University}
  \city{44th West 4th Street}
  \state{NY}
  \country{US}}
\email{pli2@stern.nyu.edu}
  
\author{Alexander Tuzhilin}
\affiliation{%
  \institution{New York University}
  \city{44th West 4th Street}
  \state{NY}
  \country{US}}
\email{atuzhili@stern.nyu.edu}
%
\begin{abstract}
Multi-criteria recommender systems have been increasingly valuable for helping consumers identify the most relevant items based on different dimensions of user experiences. However, previously proposed multi-criteria models did not take into account latent embeddings generated from user reviews, which capture latent semantic relations between users and items. To address these concerns, we utilize variational autoencoders to map user reviews into latent embeddings, which are subsequently compressed into low-dimensional discrete vectors. The resulting compressed vectors constitute latent multi-criteria ratings that we use for the recommendation purposes via standard multi-criteria recommendation methods. We show that the proposed latent multi-criteria rating approach outperforms several baselines significantly and consistently across different datasets and performance evaluation measures.

\end{abstract}

%

%
\maketitle

\section{Introduction and Related Work}
The field of recommender systems has experienced extensive growth in many research directions over the last decade \cite{adomavicius2005toward,ricci2015recommender}, including the area of multi-criteria recommendations \cite{adomavicius2015multi,sahoo2012research,jannach2012accuracy,jannach2014leveraging}. For example, several prominent websites from Zagat to TripAdvisor collect multi-criteria ratings to measure quality of items shown on their sites that can subsequently be used for recommendation purposes. 

Previous researchers try to improve accuracy of multi-criteria recommender systems in various ways, including Support Vector Regression \cite{jannach2012accuracy} or the halo effect \cite{sahoo2012research}. However, most of the methods do not take the information contained in user reviews into account, which could potentially alleviate the sparsity problem and improve quality of recommendations, as pointed out in \cite{chen2015recommender}. Some researchers propose to utilize aspect information \cite{bauman2017aspect,musto2017multi,cheng2018aspect,wang2010latent} from user reviews for multi-criteria recommendations. Although useful, these papers do not consider the latent semantic information contained in user reviews,  which is beneficial for understanding users' true experiences and expectations \cite{zhang2019deep}. Besides, user reviews contain high-dimensional user feedback information beyond aspects, while the multi-criteria ratings collected by the online platforms are limited by low-dimensional pre-defined criteria, which may not fully represent multiplicity of user experiences. Moreover, some of the multi-criteria ratings might be missing, thus limiting performance of explicit multi-criteria rating methods \cite{adomavicius2015multi}.

To address these concerns, it's natural to map the user reviews into latent embeddings and incorporate these embeddings into the recommendation process. However, typical review embeddings are noisy and high-dimensional, which significantly increases the difficulty of computation and optimization. To resolve this issue, prior literature proposed to compress text embeddings with reasonable semantic segmentation into low-dimensional discrete vectors \cite{chen2018learning,shu2017compressing}, showing that the compressed vectors manage to achieve better performance in sentiment analysis and machine translation tasks. In this paper, we extend this idea from text analysis to multi-criteria recommender systems and propose to use these ''compressed vectors'' as latent multi-criteria ratings for recommendation purposes.

Specifically, we propose to extract latent multi-criteria ratings from the user reviews using the variational autoencoder \cite{kingma2013auto}. Furthermore, we map the reviews into latent embeddings to represent high-dimensional user experiences, and then perform embedding compression (e.g., using Gumbel-Softmax Reparameterization \cite{jang2016categorical,maddison2014sampling}) to compress them into low-dimension discrete vectors, which constitute the latent multi-criteria ratings for recommendations. We empirically validate the proposed method on three datasets and demonstrate that our approach outperforms several important baselines consistently and significantly in terms of various recommendation accuracy measures.

Note that, the proposed latent multi-criteria ratings method has the following advantages over the method using multi-criteria ratings explicitly provided by the users:
\begin{itemize}
\item It is not limited by the pre-defined criteria to model user experiences.
\item It does not require to collect multi-criteria feedback from the user.
\item It does not need to deal with the missing values problem for multi-criteria ratings.
\item It captures latent interactions between users and items (using autoencoders), thus providing several benefits reported in \cite{zhang2019deep}.
\end{itemize}

In this paper, we make the following contributions. We propose a novel method that automatically generates latent multi-criteria ratings from user reviews by combining autoencoding and embedding compression techniques for multi-criteria recommendations. We also empirically demonstrate that our approach outperforms the selected baseline models consistently and significantly on three datasets across various experimental settings.

\section{Method}
\begin{figure}[h]
\centering
\includegraphics[width=0.3\textwidth]{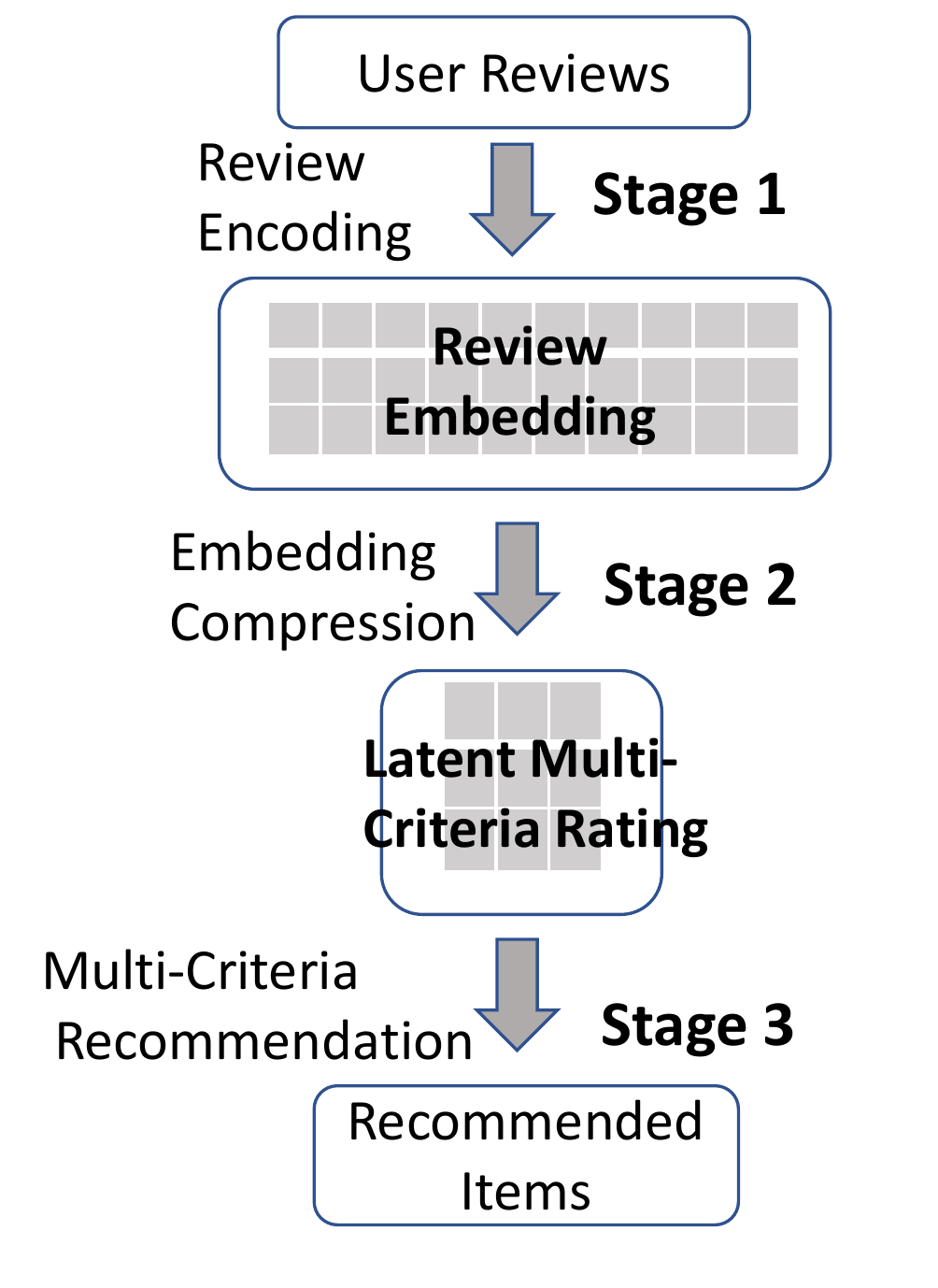}
\caption{Illustration of Our Proposed Model}
\label{process}
\end{figure}

In this section, we introduce the proposed model for latent multi-criteria recommendations by combining autoencoding and embedding compression techniques, as presented in Figure \ref{process}. In Stage 1, we use the variational autoencoder to project the user reviews onto a latent continuous space; and then we utilize the embedding compression technique to compress the embedding vectors obtained in the previous stage into discrete latent ratings during Stage 2. Finally in Stage 3, we apply various multi-criteria recommendation algorithms on the latent ratings to produce recommended items.

\subsection{Stage 1: Review Encoding}
To project the discontinuous user reviews into latent continuous embeddings, we follow the idea of autoencoding \cite{sutskever2014sequence} and implement the bidirectional GRU \cite{cho2014learning} neural networks as the encoder and the decoder respectively. Compared with classical models like RNN or LSTM, GRU is computationally more efficient and better captures semantic meanings \cite{chung2014empirical}. 

During the training process, every word $w$ in the review $s=\{w^{1},w^{2},\cdots,w^{N_{s}}\}$ is mapped to their corresponding word indexes in the pre-defined vocabulary $V$. Both the input and the output of the constructed autoencoder are the sequences of indexes. To illustrate the GRU learning procedure, we denote $[W^{z},W^{r},U^{z},U^{r}]$ as the weight matrices of current information and the past information for the update gate and the reset gate respectively. $x_{t}$ is the index vector input at the timestep $t$,  $h_{t}$ stands for the output vector, $z_{t}$ denotes the update gate status and $r_{t}$ represents the status of reset gate. The hidden state at timestep $t$ could be obtained following these equations:
\begin{equation}
z_t = \sigma_g(W_{z} x_t + U_{z} h_{t-1} + b_z) 
\end{equation}
\begin{equation}
r_t = \sigma_g(W_{r} x_t + U_{r} h_{t-1} + b_r) 
\end{equation}
\begin{equation}
h_t =  (1-z_t) \circ h_{t-1} + z_t \circ \sigma_h(W_{h} x_t + U_{h} (r_t \circ h_{t-1}) + b_h)
\end{equation}

By iteratively calculating hidden step throughout every time step, we could get the final hidden state at the end of the sentence, which constitutes the review embeddings $R = h_{N_{s}}$ that captures the latent semantic information of the user reviews. 

\subsection{Stage 2: Embedding Compression}
Note that, review embeddings obtained in Stage 1 are noisy and of high-dimensions, which significantly affect the performance of recommendation system. This motivates us to utilize embedding compression methods for efficient learning. We denote that the embedding matrix $R$ consists of $|S|$ embedding vectors with $H$ dimensions, where $|S|$ refers to the total number of reviews in the dataset. Given the arbitrary number $K$ and $M$, our goal is to find out discrete codes of dimension $K$ ($K << H$) that take integer value ranging from 1 to $M$ in the latent dimension. 

Inspired by the Gumbel-Softmax reparameterization method \cite{jang2016categorical,shu2017compressing}, we conduct the embedding compression process using the compositional code learning method that parameterizes the discrete codes onto continuous distributions. The Gumbel-Softmax technique \cite{gumbel1954statistical,maddison2014sampling} provides a simple and efficient way to get samples $z$ from a categorical distribution with class probabilities $\pi$:
\begin{equation}
z = \mbox{one-hot}(argmax_{i}[g_{i}+log \pi_{i}])
\end{equation}
where $g_{i}$ are drawn from the Gumbel distribution and $one-hot$ stands for the one-hot function that transforms the data to a binary one-hot encoding. We use the softmax function as the continuous differentiable approximation to the argmax function, so that the generated sample vectors would be:
\begin{equation} 
y_{i}=\frac{exp((log(\pi_{i})+g_{i})/\gamma)}{\sum_{K=1}^{K}exp((log(\pi_{K})+g_{K})/\gamma)}
\end{equation}
for $i=1,2,\cdots,K$ where $\gamma$ represents the temperature of the softmax function. Therefore as discussed in \cite{shu2017compressing}, if we reverse the entire sampling process described above, we can learn the discrete codes from our continuous embeddings. 

Specifically, given the number of latent dimensions $K$, we first apply the matrix factorization \cite{koren2009matrix} to get the top-K factor for that embedding matrix: $R = \sum_{i=1}^{K} D^{i}A_{i}$ where $A_{i} \in R^{K*H} $ is the basis matrix for the $i-th$ component. Therefore, the learning of discrete codes would be equivalent to the learning of a set one-hot vectors $d_{w}^{i}$ so that $ \hat{R} = \sum_{i=1}^{K}A_{i}^{T}d_{w}^{i}$ Furthermore, we assume that the discrete vectors $d_{w}^{i}$ are sampled from a prior distribution via Gumbel-Softmax Reparameterization method \cite{shu2017compressing} by minimizing the reconstruction loss compared with the origin embedding matrix,
\begin{equation}
min_{w_{i}}\frac{1}{|S|}\sum_{s\in S}||R_{s}-\hat{R_{s}}|| 
\end{equation}
where $R'_{s}$ stands for the embedding matrix reconstructed from the compressed codes we get. We optimize the prior parameters $\pi_{i}$ and $g_{i}$. Finally, the compressed vectors for each review embeddings $D_{w}$ could be obtained by applying $argmax$ to the one-hot vectors $d_{w}^{i}$.

\subsection{Stage 3: Multi-Criteria Recommendation}
When we compressed continuous review embeddings $R$ into discrete latent embeddings $D_{w}$ as described in Stage 2, we select 
the dimension of discrete codes $K$ corresponding to the dimension of collected multi-criteria ratings, and also select the range of each latent dimension $M$ matching with the range of organic multi-criteria ratings. In that sense, we could treat these compressed vectors $D_{w}$ as the \textbf{latent} multi-criteria ratings and then utilize these multi-criteria ratings for recommendation purposes, shown as Stage 3 in Figure \ref{process}. Note that, unlike traditional cases, the latent multi-criteria ratings are not specified by the users but obtained from the reviews as described in the previous stages. We conduct the multi-criteria recommendations using the state-of-the-art methods introduced in \cite{adomavicius2015multi}.

To summarize, we propose to uniquely combine the method of variational autoencoders with the embedding compression techniques to learn the latent multi-criteria ratings from user reviews for multi-criteria recommendation purposes.  In the next section, we experimentally demonstrate the superiority and effectiveness of the proposed approach.

\begin{table}[!]
\small
\centering
\begin{tabular}{|c|c|c|c|}
\hline
& \textbf{Yelp Round 8} & \textbf{Yelp Round 11} & \textbf{TripAdvisor} \\ \hline
\# Reviews & 40,314 & 11,285 & 25,928 \\ 
\# Items & 1,134 & 7,788 & 2,933 \\
\# Users & 600 & 587 & 4,252 \\
Sparsity & 5.89\% & 0.25\% & 0.21\% \\ \hline
\end{tabular}
\newline
\caption{Descriptive Statistics of Three Datasets}
\label{statisticalnumber}
\end{table}

\begin{table*}[!]
\footnotesize
\centering
\begin{tabular}{|c|c|cccc|cccc|cccc|} \hline 
    \multirow{2}{*}{Algorithm} & \multirow{2}{*}{Rating} & \multicolumn{4}{c|}{Yelp Round 11} & \multicolumn{4}{c|}{Yelp Round 8} & \multicolumn{4}{c|}{TripAdvisor} \\ \cline{3-14}
    & & Pre@1 & Pre@5 & Rec@1 & Rec@5 & Pre@1 & Pre@5 & Rec@1 & Rec@5 & Pre@1 & Pre@5 & Rec@1 & Rec@5\\ \hline
    \multirow{5}{*}{KNN} & \textbf{LatentMC}   &  \textbf{0.8150*} & \textbf{0.6822} & \textbf{0.2401*} & \textbf{0.8387*} & \textbf{0.7275*} & \textbf{0.6864} & \textbf{0.3084*} & \textbf{0.6907*} & \textbf{0.7568*} & \textbf{0.6792} & \textbf{0.3472*} & \textbf{0.7655*} \\
                              & MC & 0.7743 & 0.6775 & 0.2199 & 0.8270 & 0.7092 & 0.6741 & 0.2880 & 0.6772 & 0.7298 & 0.6529 & 0.3208 & 0.7536 \\
                              & Overall & 0.7920 & 0.6750 & 0.2327 & 0.8272 & 0.7078 & 0.6543 & 0.2902 & 0.6659 & 0.7304 & 0.6609 & 0.3232 & 0.7544 \\
                              & Embedding & 0.7920 & 0.6720 & 0.2309 & 0.8213 & 0.6978 & 0.6582 & 0.2856 & 0.6709 & 0.7312 & 0.6718 & 0.3220 & 0.7088 \\
                              & PCA & 0.7918 & 0.6704 & 0.2268 & 0.8000 & 0.7048 & 0.6602 & 0.2875 & 0.6660 & 0.7314 & 0.6788 & 0.3199 & 0.7332 \\
                              & Aspect & 0.7977 & 0.6768 & 0.2315 & 0.8305 & 0.7056 & 0.6624 & 0.2844 & 0.6718 & 0.7285 & 0.6042 & 0.3244 & 0.7620 \\ \hline
    \multirow{5}{*}{SlopeOne} & \textbf{LatentMC}  & \textbf{0.8400*} & \textbf{0.6953*} & \textbf{0.2382*} & \textbf{0.8395*} & \textbf{0.7268*} & \textbf{0.6947*} & \textbf{0.3028*} & \textbf{0.6890} & \textbf{0.7563*} & \textbf{0.6785*} & \textbf{0.3458*} & \textbf{0.7675} \\
                              & MC & 0.7760 & 0.6777 & 0.2242 & 0.8254 & 0.7078 & 0.6686 & 0.2903 & 0.6742 & 0.7308 & 0.6666 & 0.3276 & 0.7076 \\
                              & Overall & 0.7920 & 0.6714 & 0.2258 & 0.8210 & 0.7066 & 0.6636 & 0.2912 & 0.6744 & 0.7296 & 0.6624 & 0.3250 & 0.7067 \\
                              & Embedding & 0.7663 & 0.6819 & 0.2004 & 0.8253 & 0.7092 & 0.6820 & 0.2880 & 0.6836 & 0.7270 & 0.6066 & 0.3228 & 0.5906 \\
                              & PCA & 0.7992 & 0.6705 & 0.2205 & 0.8020 & 0.7078 & 0.6692 & 0.2793 & 0.6620 & 0.7195 & 0.6264 & 0.3220 & 0.6880 \\
                              & Aspect & 0.7600 & 0.6704 & 0.2088 & 0.8204 & 0.7003 & 0.6838 & 0.2858 & 0.6836 & 0.7285 & 0.6042 & 0.3244 & 0.7620 \\ \hline
    \multirow{5}{*}{CoCluster} & \textbf{LatentMC}   & \textbf{0.8400*} & \textbf{0.6880*} & \textbf{0.2414*} & \textbf{0.8379*} & \textbf{0.7292*} & \textbf{0.6896*} & \textbf{0.3030*} & \textbf{0.6930*} & \textbf{0.7550*} & \textbf{0.6566} & \textbf{0.3468*} & \textbf{0.7815*} \\
                              & MC & 0.8160 & 0.6688 & 0.2295 & 0.8251 & 0.7068 & 0.6642 & 0.2928 & 0.6730 & 0.7332 & 0.6510 & 0.3280 & 0.7628 \\
                              & Overall & 0.7520 & 0.6736 & 0.2173 & 0.8221 & 0.7076 & 0.6795 & 0.2920 & 0.6830 & 0.7318 & 0.6436 & 0.3288 & 0.7355 \\
                              & Embedding & 0.8080 & 0.6784 & 0.2334 & 0.8290 & 0.7024 & 0.6695 & 0.2910 & 0.6795 & 0.7296 & 0.6548 & 0.3302 & 0.6476 \\
                              & PCA & 0.8114 & 0.6759 & 0.2119 & 0.8000 & 0.7078 & 0.6692 & 0.2793 & 0.6620 & 0.7190 & 0.6464 & 0.3302 & 0.6476 \\
                              & Aspect & 0.7823 & 0.6624 & 0.2284 & 0.8214 & 0.6998 & 0.6686 & 0.2902 & 0.6788 & 0.7325 & 0.6166 & 0.3268 & 0.7229 \\ \hline
    \multirow{5}{*}{SVR} & \textbf{LatentMC}   & \textbf{0.8792*} & \textbf{0.7077*} & \textbf{0.2984*} & \textbf{0.8808*} & \textbf{0.7868*} & \textbf{0.7890*} & \textbf{0.3348*} & \textbf{0.9432} & \textbf{0.8382*} & \textbf{0.9774*} & \textbf{0.3692*} & \textbf{0.5392*} \\
                              & MC & 0.8568 & 0.6894 & 0.2802 & 0.8656 & 0.7692 & 0.7778 & 0.3130 & 0.9333 & 0.8072 & 0.9502 & 0.3380 & 0.5210 \\
                              & Overall & 0.8608 & 0.6880 & 0.2810 & 0.8670 & 0.7538 & 0.7767 & 0.3098 & 0.9255 & 0.8002 & 0.9530 & 0.3298 & 0.5210 \\
                              & Embedding & 0.8590 & 0.6866 & 0.2798 & 0.8656 & 0.7550 & 0.7680 & 0.3104 & 0.9020 & 0.7988 & 0.9530 & 0.3336 & 0.5310 \\
                              & PCA & 0.8392 & 0.6820 & 0.2798 & 0.8650 & 0.7566 & 0.7660 & 0.3100 & 0.8998 & 0.8002 & 0.9459 & 0.3330 & 0.5310 \\
                              & Aspect & 0.8608 & 0.6902 & 0.2815 & 0.8700 & 0.7593 & 0.7756 & 0.3098 & 0.9333 & 0.8036 & 0.9414 & 0.3358 & 0.5175 \\ \hline
    \multirow{5}{*}{Aggregate} & \textbf{LatentMC}   & \textbf{0.8230*} & \textbf{0.6928*} & \textbf{0.2478*} & \textbf{0.8400*} & \textbf{0.7532*} & \textbf{0.7578*} & \textbf{0.3128*} & \textbf{0.6988*} & \textbf{0.6978*} & \textbf{0.4073*} & \textbf{0.3532*} & \textbf{0.7001*} \\
                              & MC & 0.8088 & 0.6808 & 0.2302 & 0.8292 & 0.7388 & 0.7402 & 0.2978 & 0.6782 & 0.6767 & 0.3819 & 0.3318 & 0.6862 \\
                              & Overall & 0.8072 & 0.6792 & 0.2315 & 0.8178 & 0.7388 & 0.7440 & 0.2956 & 0.6664 & 0.6780 & 0.3270 & 0.3306 & 0.6171 \\
                              & Embedding & 0.8028 & 0.6820 & 0.2328 & 0.8230 & 0.7402 & 0.7398 & 0.3002 & 0.6792 & 0.6767 & 0.3335 & 0.3298 & 0.6363 \\
                              & PCA & 0.7978 & 0.6780 & 0.2315 & 0.8230 & 0.7398 & 0.7398 & 0.2988 & 0.6660 & 0.6767 & 0.3330 & 0.3306 & 0.6360 \\
                              & Aspect & 0.8078 & 0.6808 & 0.2318 & 0.8208 & 0.7356 & 0.7402 & 0.2972 & 0.6848 & 0.6802 & 0.3875 & 0.3320 & 0.6787 \\ \hline
\end{tabular}
\newline
\caption{Experiment results on three datasets. * stands for significance under 95\% confidence}
\label{result}
\end{table*}

\section{Experiments and Results}
\subsection{Experimental Settings}
We implement the model on the following datasets: Yelp Challenge Dataset \footnote{https://www.yelp.com/dataset} Round 8 and Round 11 of restaurant recommendations and on TripAdvisor Dataset \footnote{http://www.cs.cmu.edu/˜jiweil/html/hotel-review.html} of hotel recommendations that contain user reviews, as well as multi-criteria feedback. The difference between two Yelp datasets is the time when the data is collected: Round 8 dataset is collected in 2016, while Round 11 dataset is collected in 2018. To avoid the sparsity and the cold start problems, we only consider users that rate at least five items, restaurants that have been rated by at least ten users and hotels that have been rated by at least five users. The basic statistics of the filtered datasets are shown in Table \ref{statisticalnumber}. (Note that the unfiltered datasets are much larger.)

To demonstrate the effectiveness of the proposed latent multi-criteria rating (LatentMC) model, we select several multi-criteria rating models for comparison, including:
\begin{itemize}
\item \textbf{MC}: the multi-criteria ratings explicitly specified by the users in the three datasets.
\item \textbf{Overall}: the overall ratings in the three dataset.
\item \textbf{Embedding}: the uncompressed review embeddings obtained in Stage 1 (see Figure 1) as latent multi-criteria ratings in the three datasets.
\item \textbf{Aspect}: the extracted aspect ratings \cite{bauman2017aspect} as multi-criteria ratings in the three datasets.
\item \textbf{PCA}: the top-k factors extracted using Principal Component Analysis \cite{jolliffe2011principal} as latent multi-criteria ratings in the three datasets.
\end{itemize}

Also, to evaluate recommendation performance, we implement the state-of-the-art multi-criteria recommendation models introduced in \cite{adomavicius2015multi}, including KNN, SlopeOne, CoCluster, Support Vector Regression (SVR) and Aggregate Function. The first three models are adjusted to their multi-criteria recommendation versions by calculating similarities using multi-criteria ratings instead of only overall ratings.

\subsection{Experimental Results}
We conduct the 5-fold cross-validation recommendation experiments and report the performance results on the test set. As shown in Table \ref{result}, our proposed LatentMC model outperforms the baselines consistently and significantly across different datasets, performance measures and algorithms.  For example using Aggregate algorithm, the Pre@1, Pre@5, Rec@1, Rec@5 measures for the three datasets improve by 1.73\%, 1.56\%, 6.05\%, 1.29\%, 1.91\%, 2.32\%, 4.03\%, 2.00\%, 2.52\%, 4.86\%, 5.47\% and 1.99\% respectively compared to the second-best baseline models. In general, we obtain over 2\% accuracy improvements in most of the cases.

Furthermore, we make the following observations. First, compared with the recommendation results directly using uncompressed review embeddings, the latent compressed multi-criteria rating method achieves better performance. The improvement is achieved through the compression process that cleans up the noisy high-dimensional embedding vectors and extracts the essence of the reviews. Second, we observe that review-based multi-criteria recommendation model performs better that the non-review recommendation model, even without using the explicit multi-criteria rating information. This indicates that user reviews contain richer and high-dimensional information, compared to low-dimensional multi-criteria ratings. Thus, it is crucial to properly model user reviews in the recommendation process. Finally, the latent rating method outperforms the explicit rating models because they capture the latent semantic information within user reviews, which supports the general advantages of using latent methods in text analysis \cite{zhang2019deep}.

\vspace{-0.5cm}
\subsection{Conclusions}
To conclude, the latent multi-criteria ratings generated by the proposed model achieve significantly better performance for multi-criteria recommendations in comparison to the alternative methods. In addition, it has the following natural advantages over classical multi-criteria recommendation approaches. First, it is not limited by the pre-defined criteria and missing values to model user experiences. Second, it does not require to collect multi-criteria feedback from the user. Third, it captures latent interactions between users and items, which provide several benefits reported in \cite{zhang2019deep}.

As the future work, we would like to improve the latent multi-criteria rating generation process even further. Also, we plan to study the semantic meaning and interpretability of the latent multi-criteria ratings.

\bibliographystyle{ACM-Reference-Format}
\bibliography{sigproc}

\end{document}